\documentclass[letterpaper, 10pt, conference]{ieeeconf}  

\let\labelindent\relax
\usepackage{microtype} 
\usepackage{booktabs}
\usepackage{multirow}
\usepackage{multicol} 
\usepackage{xcolor}
\usepackage{colortbl} 
\usepackage{setspace} 
\usepackage{mathtools} 
\usepackage{pifont}       
\usepackage{bbding}       
\usepackage{fontawesome}  

\IEEEoverridecommandlockouts

\usepackage{graphics} 
\usepackage{graphicx} 
\usepackage{amsmath} 
\usepackage{amssymb}  
\usepackage{cite}
\makeatletter
\let\NAT@parse\undefined
\makeatother
\usepackage[colorlinks=true, citecolor=orange, linkcolor=magenta, urlcolor=cyan]{hyperref}
\usepackage{enumitem}

\title{\LARGE \bf
\textit{VLN-Cache}: Enabling Token Caching for VLN Models with \\ Visual/Semantic Dynamics Awareness
}

\author{$^{*}$Zihao Zheng$^{1}$, $^{*}$Zhihao Mao$^{2}$, Xingyue Zhou$^{3}$, Jiayu Chen$^{1}$, Maoliang Li$^{1}$, Xinhao Sun$^{4}$, Hailong Zou$^{1}$, \\ Zhaobo Zhang$^{1}$, Xuanzhe Liu$^{1}$, Donggang Cao$^{1}$, Hong Mei$^{1}$,$^{\dagger}$ Xiang Chen$^{1}$
\thanks{$^{*}$ Equal Contribution}
\thanks{$^{1}$ School of Computer Science, Peking University, Beijing, China.}
\thanks{$^{2}$ School of Computer Science, China University of Geosciences (Wuhan), Wuhan, China.}
\thanks{$^{3}$ School of Artificial Intelligence and Automation, Huazhong University of Science and Technology, Wuhan, China.}
\thanks{$^{4}$ School of Electronics Engineering and Computer Science, Peking University, Beijing, China.}
\thanks{$^{\dagger}$ Corresponding Author: Xiang Chen$<${\tt\small xiang.chen@pku.edu.cn}$>$}
}

\begin{document}

\maketitle
\thispagestyle{empty}
\pagestyle{empty}

\begin{abstract}

Vision-and-Language Navigation (VLN) increasingly relies on large vision-language models, but their inference cost conflicts with real-time deployment.
Token caching is a promising training-free strategy that avoids redundant computation by reusing stable visual tokens across frames.
However, most existing token caching methods are not designed for embodied navigation, where continuous viewpoint changes and instruction-conditioned task-stage shifts jointly affect safe reuse.
We identify two failure modes: (1) visual dynamics, where viewpoint shift displaces token positions across frames, causing position-wise matching to pair misaligned content; (2) semantic dynamics, where token relevance shifts across task stages as navigation progresses, making cached states stale.
We propose \textit{VLN-Cache}, a visual-dynamic-aware and semantic-dynamic-aware caching framework that introduces view-aligned remapping to recover geometric correspondences and a task-relevance saliency filter to veto reuse at semantic transitions.
A layer-adaptive entropy policy further balances the per-layer reuse budget.
Experiments on the R2R-CE simulation benchmark show up to 1.52$\times$ speedup while maintaining competitive navigation success rates.

\end{abstract}

\section{\textbf{Introduction}}
\label{sec:introduction}

Vision-and-Language Navigation (VLN) enables embodied agents to follow natural language instructions in complex, unstructured environments~\cite{VLN_servey1, VLN_servey2}.
It has become a mainstream paradigm in the field of embodied intelligence~\cite{VL_Nav, IROS_VLN}.
Modern high-performance VLN planners are built on large VLMs~\cite{RT2, OpenVLA} that must perform a full forward pass at every navigation step, making per-step latency a critical bottleneck for real-time robotic deployment~\cite{HIAI}.

\begin{figure*}[!t]
\centering
\includegraphics[width=\textwidth]{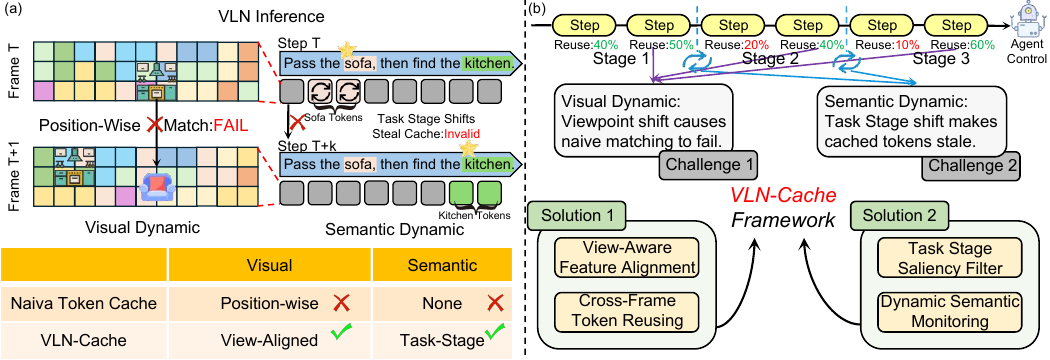}
\caption{Overview of the Proposed \textit{VLN-Cache} Framework.}
\vspace{-4mm}
\label{fig:analysis-overview}
\end{figure*}

Existing acceleration approaches for VLN primarily follow three directions: efficient architectures, model compression, and runtime optimization.
Efficient architectures design lightweight networks to reduce inherent complexity~\cite{efficientvln,etp_r1}.
Model-level efficiency methods reduce cost through distillation or hierarchical action abstraction~\cite{MiniVLN, NaVILA}.
Runtime optimization dynamically reduces computations during inference without altering model parameters~\cite{NAPVLN, Uni-NaVid}.
These methods have substantially improved VLN inference efficiency.

Token caching has emerged as a highly promising training-free runtime optimization technique.
It exploits temporal coherence: background regions such as walls and floors change little across adjacent frames, so their visual tokens can be safely reused rather than recomputed~\cite{vlacache,stcache, token_sparsification, EgoPrune}.
Following a detect-and-reuse paradigm, existing methods selectively skip the computation of tokens identified as static or task-irrelevant, achieving substantial speedup with minimal accuracy loss in fixed-camera or static-scene settings~\cite{vlacache, token_merging, stcache, EgoPrune, token_sparsification, Learn_VLA_cache}.

However, existing token caching methods are built on a static-view assumption: patches at the same image position are expected to remain similar across adjacent frames~\cite{vlacache}.
This assumption breaks in VLN, where the agent continuously translates and rotates during navigation, causing physically static objects to shift substantially in image coordinates and making position-wise token matching unreliable~\cite{View_Invariant_Learning}.
In addition, as the agent executes the instruction step by step, the semantic relevance of many image patches changes along the trajectory.
A landmark that is critical before a turn—guiding the agent toward a decision point—can quickly become irrelevant once the agent has passed it, even though its visual appearance remains largely unchanged.
These two dynamics jointly limit the direct transfer of static-scene caching strategies to embodied navigation without adaptation.

To investigate these dynamics, we conduct an empirical analysis of visual dynamics and semantic dynamics across VLN trajectories.
For visual dynamics, we analyze how viewpoint changes affect cross-frame token consistency.
For semantic dynamics, we analyze how task progression changes region relevance over time.
Our analysis reveals two consistent findings: view-aligned matching achieves on average ${\sim}10.3\%$ higher cross-frame token similarity than position-wise matching, directly recovering reuse opportunities lost to viewpoint shift; and the task relevance of landmark regions varies substantially across a trajectory even when appearance is stable, motivating a dynamic semantic monitoring mechanism.

Motivated by these findings, we propose \textit{VLN-Cache}, a dual-aware token caching framework for VLN.
For visual dynamics, we perform view-aligned remapping before cross-frame matching, which recovers reusable tokens that would be missed by position-wise comparison.
For semantic dynamics, we introduce a task-relevance saliency filter that refreshes cached tokens when their task relevance changes.
To control the additional overhead, we further adopt a layer-adaptive entropy policy that allocates conservative reuse to sensitive layers and more aggressive reuse to stable layers.

Our key contributions are as follows:
\begin{itemize}[leftmargin=*]
    \item[$\bullet$] We provide empirical evidence that static-scene caching assumptions fail in VLN due to viewpoint-induced mismatch and temporal semantic shift.
    \item[$\bullet$] We present \textit{VLN-Cache}, a dual-aware framework that combines view-aware matching with task-relevance semantic refresh, without architectural changes or retraining.
    \item[$\bullet$] We design an entropy-based layer-adaptive reuse strategy to balance acceleration gain and computational overhead across transformer layers.
\end{itemize}

Experimental results demonstrate that \textit{VLN-Cache} achieves a speedup of up to $1.52\times$ on standard VLN benchmarks, while maintaining competitive navigation success rates, supporting practical real-time embodied deployment.
\section{\textbf{Background}}
\label{sec:background}

\subsection{\textbf{Efficient Vision-Language Navigation}}

Large VLA models have driven significant accuracy gains in VLN~\cite{VLN_servey1, VLN_servey2, VL_Nav, IROS_VLN}, but their per-step inference cost makes real-time deployment difficult.
Existing efficiency work broadly splits into three directions.
The first designs lightweight architectures: EfficientVLN~\cite{efficientvln} uses progressive recursive memory, and ETP~\cite{etp_r1} combines topological planning with RL fine-tuning.
The second compresses the model: MiniVLN~\cite{MiniVLN} uses distillation, and NaVILA~\cite{NaVILA} adopts a two-tier VLA-locomotion policy.
The third reduces runtime computation without modifying the model: Uni-NaVid~\cite{Uni-NaVid} compresses temporal observation streams, and KERV~\cite{KERV} rectifies kinematic bias for speculative decoding in embodied VLA inference.
These methods treat each step independently and do not exploit redundancy across consecutive frames, leaving cross-frame token reuse unexplored.
Our work operates orthogonally as a training-free inference-time caching wrapper.

\subsection{\textbf{Token Caching and Sparsification in VLA}}

Selective KV cache compression has proven effective in LLM decoding~\cite{SnapKV}; token caching extends this idea to the visual domain by leveraging the high inter-frame similarity in sequential observations to skip re-computation for stable tokens.
In the VLA setting, VLACache~\cite{vlacache} and Learn-VLA-Cache~\cite{Learn_VLA_cache} identify temporally redundant tokens and reuse their cached states across frames.
A parallel line prunes or merges tokens outright: FastV~\cite{FastV} and LLaVA-PruMerge~\cite{LLaVA_PruMerge} drop or merge inattentive visual tokens in VLMs; in VLA pipelines, STCache~\cite{stcache} drops spatio-temporal tokens via semantic-aware accumulation, token merging~\cite{token_merging} aggregates near-duplicate tokens via bipartite matching, and EgoPrune~\cite{EgoPrune} uses perspective-aware, egomotion-aligned token selection.
Despite their effectiveness in static or manipulation settings, most rely on local appearance similarity and lack VLN-specific geometry or task-stage modeling, implicitly assuming the same patch coordinate in frame $t$ observes the same physical content as in frame $t{-}1$.
In VLN, continuous viewpoint shift displaces scene content across patch coordinates, so position-wise matching pairs misaligned tokens and fail silently; a low similarity score may reflect misalignment rather than genuine scene change.
Sec.~\ref{sec:analysis} quantifies this gap directly on VLN trajectories and motivates our dual-aware design.

\section{\textbf{Visual/Semantic Dynamics Analysis}}
\label{sec:analysis}

We identify two independent dynamics in VLN inference that jointly undermine the effectiveness of position-wise token caching: visual dynamics arising from continuous viewpoint shift and semantic dynamics arising from task progression.

\subsection{\textbf{Visual Dynamics from Viewpoint Shift}}
\label{sec:vis-dynamics}

\noindent \textit{Motivation \ding{172}: Position-wise token caching assumes the same patch coordinate corresponds to the same physical content across frames, but this breaks down in VLN where continuous agent rotation shifts every token's image coordinate.}

Token caching assumes that adjacent frames are nearly identical, so tokens at the same spatial index represent the same physical content.
This holds reasonably well for static-camera VQA or manipulation tasks where the camera barely moves.
VLN is fundamentally different: the agent continuously translates and rotates through unseen 3D environments, so the same patch coordinate in two consecutive frames may correspond to entirely different surfaces.
A corridor wall that occupies patches $[i, i+k]$ in frame $t{-}1$ may shift to $[i+\delta, i+k+\delta]$ in frame $t$ after a rotation, and position-wise matching silently pairs the wrong content.
This systematic misalignment causes two compounding errors: (1) reused tokens carry stale content from the wrong surface, introducing noise into attention computation; (2) truly reusable tokens that have merely shifted coordinates are incorrectly marked for re-computation, wasting inference budget.

To quantify this effect, we measure two cross-frame similarity scores for each token $i$:
\begin{equation}
r^{\mathrm{pos}}_{t,i}=\cos\!\left(v_t^{(i)},v_{t-1}^{(i)}\right), \quad
r^{\mathrm{align}}_{t,i}=\cos\!\left(v_t^{(i)},v_{t-1}^{(\pi_t(i))}\right),
\end{equation}
where $v_t^{(i)}$ is the visual feature of token $i$ at step $t$, and $\pi_t(i)$ is the view-aligned correspondence obtained from depth and relative camera pose.
The former compares tokens at the same position, while the latter compares tokens that depict the same physical point.
We summarize the gap between them as the reuse gap:
\begin{equation}
\Delta r_t=\frac{1}{M}\sum_{i=1}^{M}\left(r^{\mathrm{align}}_{t,i}-r^{\mathrm{pos}}_{t,i}\right),
\label{eq:delta-r}
\end{equation}
which directly measures the reuse potential lost by position-only matching.
A positive $\Delta r_t$ means that view-aligned matching identifies more reusable tokens than position-wise matching: tokens that position-wise caching would either wrongly reuse (mismatched pair, high noise) or wrongly refresh (displaced but stable content, wasted compute).

\begin{figure}[!t]
    \centering
    \includegraphics[width=\columnwidth]{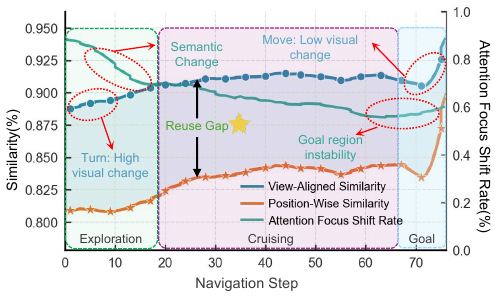}
    \caption{Visual and semantic dynamics along a VLN trajectory.}
    \vspace{-4mm}
    \label{fig:analyze-1}
\end{figure}

\begin{figure}[!t]
    \centering
    \includegraphics[width=\columnwidth]{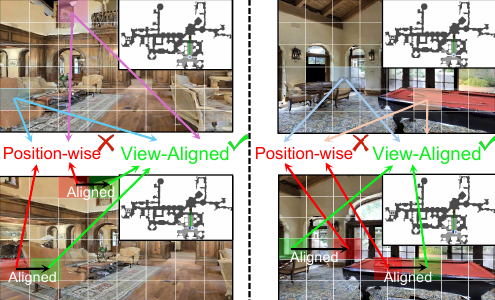}
    \caption{Position-wise token mismatch under viewpoint shift.}
    \vspace{-4mm}
    \label{fig:analyze-2}
\end{figure}

As shown in Fig.~\ref{fig:analyze-1}, $\Delta r_t$ remains persistently positive throughout navigation, averaging ${\sim}10.3\%$ across trajectories.
The gap is especially pronounced during the \textit{Exploration} phase (frequent turns, high viewpoint change) and the \textit{Goal} phase (close-range approach with abrupt viewpoint shifts), and relatively lower during the \textit{Cruising} phase (mostly forward motion).
Fig.~\ref{fig:analyze-2} illustrates a concrete mismatch case: a rotation causes a static wall region to drift four patch positions, and position-wise matching pairs it with a different surface.
These observations confirm that view-aware alignment is a necessary precondition for any reliable reuse decision in VLN.

\noindent \textit{Insight \ding{172}: The persistently positive $\Delta r_t$ shows that viewpoint shift displaces rather than destroys reusability; what fails is position-wise matching, not the reuse opportunity itself, motivating view-aligned remapping.}

\subsection{\textbf{Semantic Dynamics from Task Stage}}
\label{sec:sem-dynamics}

\noindent \textit{Motivation \ding{173}: View-aligned remapping resolves geometric misalignment but is blind to task progression, where navigational relevance shifts as sub-goals are completed, making visually stable tokens semantically stale.}

Visual dynamics address \emph{where} tokens should be matched across frames; semantic dynamics address \emph{whether} a token should be reused at all, regardless of its geometric stability.
Even when two tokens are correctly aligned by view-aware correspondence and visually similar, reusing cached states can still be harmful if the semantic role of that region has shifted.
During navigation, the agent progressively follows a multi-step instruction: a hallway that guided early waypoint decisions becomes irrelevant once passed, while a previously background object becomes the next landmark.
This shift is task-driven rather than appearance-driven: the visual content of the hallway does not change, but its relevance to the current sub-goal drops sharply once the corresponding instruction segment is completed.
Tokens in these transitioning regions carry fresh decision-critical information that a stale cached entry from even one step ago cannot faithfully represent.
Reusing such states risks propagating outdated attention patterns into the model's action prediction, which is particularly harmful at sub-goal boundaries where the agent must reorient its focus.
Unlike visual dynamics, which can be resolved by geometric alignment alone, semantic dynamics require explicit and continuous monitoring of instruction-conditioned relevance as the agent progresses.

To characterize this phenomenon, we track the set of top-$k$ vision tokens with the highest instruction-conditioned attention scores at each step.
We measure the shift between adjacent steps as the Jaccard distance over these focus sets:
\begin{equation}
D_t^{\mathrm{sem}}=1-\frac{|\mathcal{S}_t\cap\mathcal{S}_{t-1}|}{|\mathcal{S}_t\cup\mathcal{S}_{t-1}|},
\label{eq:sem-jd}
\end{equation}
where $\mathcal{S}_t$ is the top-$k$ attended token index set at step $t$.
$D_t^{\mathrm{sem}}=0$ means the attended region is unchanged; $D_t^{\mathrm{sem}}=1$ means a complete turnover.
A high value indicates that task focus has shifted to an entirely different set of visual regions, making previously cached token states semantically stale.

As shown in Fig.~\ref{fig:analyze-1} (right axis), the attention focus shift rate is high during the Exploration phase, when the agent is rapidly scanning the environment and task focus is volatile, and decreases during Cruising, when the agent follows a stable path segment.
This temporal pattern confirms that semantic dynamics are persistent but non-uniform; they intensify at navigation phase boundaries and cannot be captured by any geometry-only reuse criterion.

\noindent \textit{Insight \ding{173}: The substantial and phase-varying $D_t^{\mathrm{sem}}$ reveals an orthogonal failure mode: a token can be visually stable yet semantically stale, motivating a task-relevance gate that vetoes reuse when task focus shifts.}

These two dynamics constitute independent failure modes that existing caching methods are not designed to handle, motivating the dual-aware framework in Sec.~\ref{sec:methods}.


\section{\textbf{\textit{VLN-Cache} Framework}}
\label{sec:methods}

\begin{figure}[!t]
    \centering
    \includegraphics[width=\columnwidth]{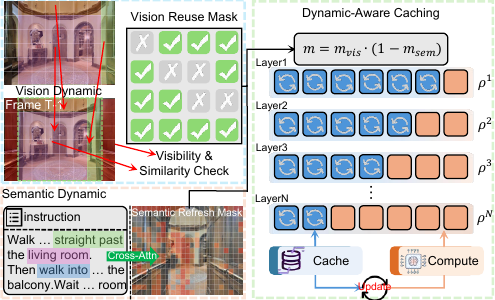}
    \caption{Overview of the VLN-Cache framework.}
    \vspace{-4mm}
    \label{fig:method-framework}
\end{figure}

Motivated by the two dynamics identified in Sec.~\ref{sec:analysis}, we propose \textit{VLN-Cache}, a training-free inference-time framework that makes visually and semantically dual-aware reuse decisions at each navigation step (Fig.~\ref{fig:method-framework}).

\subsection{\textbf{Visual-Dynamic-Aware Token Caching Scheme}}
\label{sec:vis-cache}

\subsubsection{\textbf{Visual-Dynamics Identification}}
Because the agent's camera continuously moves, a token at position~$i$ in frame~$t$ generally does not correspond to the same physical surface as position~$i$ in frame~$t{-}1$.
Existing methods match tokens purely by spatial index, causing systematic misalignment (Sec.~\ref{sec:vis-dynamics}).
We resolve this by view-aligned remapping: for each token center~$u_t^{(i)}$ with its associated depth~$d_t^{(i)}$, we back-project to 3D via camera intrinsics~$\mathbf{K}$, apply the relative pose transform $\mathbf{T}_{t\to t{-}1}$, and re-project onto the image plane to obtain its correspondence in the previous frame,
\begin{equation}
\pi_t(i)=\mathcal{N}\!\bigl(\Pi\!\bigl(\mathbf{T}_{t\to t{-}1}\,\Pi^{-1}\!\bigl(u_t^{(i)},d_t^{(i)};\mathbf{K}\bigr)\bigr)\bigr),
\label{eq:view-remap}
\end{equation}
where $\Pi^{-1}(\cdot)$ and $\Pi(\cdot)$ denote back-projection and projection, and $\mathcal{N}(\cdot)$ applies $3{\times}3$ neighborhood refinement to handle discretization mismatch between continuous projected coordinates and discrete patch indices.
A successful geometric projection alone does not guarantee safe reuse: the underlying surface may have changed due to occlusion, lighting variation, or dynamic scene content.
We therefore apply a two-stage gate: a token is marked as visually reusable ($m_{\mathrm{vis},t}^{(i)}{=}1$) only if (1)~$\pi_t(i)$ falls within the valid field of view, ruling out tokens whose corresponding surface has moved out of frame, and (2)~$\cos(v_t^{(i)},\,v_{t-1}^{(\pi_t(i))})>\tau_{\mathrm{vis}}$, confirming that the remapped pair genuinely depicts the same physical content.

\subsubsection{\textbf{Visual-Dynamics Utilization}}
For a token with $m_{\mathrm{vis},t}^{(i)}{=}1$, we retrieve cached states from the geometrically aligned position $\pi_t(i)$ rather than from the same-index position~$i$.
This shift in retrieval index is the key departure from position-wise caching: instead of asking ``is token $i$ similar to the old token $i$'' (a comparison that pairs misaligned content), we ask ``is token $i$ similar to the token that actually observed the same surface?''
Tokens that fall outside the valid field of view or lack reliable depth estimates are excluded from the reuse set and always refreshed, since no valid aligned counterpart exists in the cache.
The actual cache update mechanism using $\pi_t(i)$ is formalized in Sec.~\ref{sec:fusion}.

\subsection{\textbf{Semantic-Dynamic-Aware Token Caching Scheme}}
\label{sec:sem-cache}

\subsubsection{\textbf{Semantic-Dynamics Identification}}
Even when two tokens are geometrically aligned and visually similar, reusing cached states is harmful if the semantic role of that region has shifted.
As discussed in Sec.~\ref{sec:sem-dynamics}, the agent's instruction progressively redirects attention: a hallway that guided early navigation becomes irrelevant once passed, while a background object becomes the next landmark.
To detect such transitions, we compute an instruction-conditioned relevance score~$s_t^{(i)}$ for each token at every step.
A token is flagged for mandatory refresh ($m_{\mathrm{sem},t}^{(i)}{=}1$) under either condition: high current relevance ($s_t^{(i)}{>}\tau_{\mathrm{abs}}$), meaning it carries strong task signal a stale cached version cannot represent, or rapid relevance change ($|s_t^{(i)}{-}s_{t-1}^{(i)}|{>}\tau_{\Delta}$), indicating an ongoing semantic transition regardless of current magnitude.

\subsubsection{\textbf{Semantic-Dynamics Utilization}}
The semantic refresh mask $m_{\mathrm{sem}}$ functions as a hard veto: any token satisfying either detection condition is unconditionally routed to full re-computation, regardless of its geometric stability.
This asymmetry is intentional. Visual stability is a necessary but insufficient condition for safe reuse; a token can be perfectly aligned across frames yet carry stale task signal if the instruction focus has shifted.
By treating semantic dynamics as a veto rather than a soft penalty, VLN-Cache avoids the error mode where a high-confidence visual match silences a semantically important refresh.
The two detection thresholds $\tau_{\mathrm{abs}}$ and $\tau_{\Delta}$ govern the strictness of the veto: tighter thresholds flag more tokens for refresh (higher accuracy, lower reuse ratio), while looser thresholds allow more caching (higher speedup, more semantic staleness risk).
In practice, we jointly calibrate them on a small held-out trajectory set and report a sensitivity analysis in Sec.~\ref{sec:experiments}.
The veto is applied in the unified mask construction of Sec.~\ref{sec:fusion}.

\subsection{\textbf{Dual-Aware Fusion and Caching Policy}}
\label{sec:fusion}

\subsubsection{\textbf{Reuse Mask}}
The two identification signals are merged multiplicatively:
\begin{equation}
m_t^{(i)}=m_{\mathrm{vis},t}^{(i)}\cdot\bigl(1-m_{\mathrm{sem},t}^{(i)}\bigr).
\label{eq:reuse-mask}
\end{equation}
A token is reused only when it is geometrically stable \emph{and} not undergoing a semantic transition; either condition alone does not suffice.
The multiplicative form encodes strict AND semantics: visual stability is necessary but not sufficient, and a semantically stale token is always refreshed even when its geometry is stable.

\subsubsection{\textbf{Cross-Frame Token Splice}}
Given the reuse mask, cached token states at each decoder layer~$\ell$ are updated as:
\begin{equation}
\begin{aligned}
\hat{K}_{t}^{\ell,(i)} &= m_t^{(i)}\, K_{t-1}^{\ell,(\pi_t(i))} + \bigl(1{-}m_t^{(i)}\bigr)\, K_t^{\ell,(i)}, \\
\hat{V}_{t}^{\ell,(i)} &= m_t^{(i)}\, V_{t-1}^{\ell,(\pi_t(i))} + \bigl(1{-}m_t^{(i)}\bigr)\, V_t^{\ell,(i)}.
\end{aligned}
\label{eq:token-splice}
\end{equation}
Reused tokens are fetched from the view-aligned cache at $\pi_t(i)$; refreshed tokens receive exact model computation.
Every token position is filled at each step (from cache or fresh compute), avoiding the irregular tensor shapes and attention masking overhead of hard token dropping.
VLN-Cache wraps the model's forward pass without modifying weights or attention kernels, making it a drop-in wrapper for any transformer-based VLA.

\subsubsection{\textbf{Layer-Adaptive Caching Policy}}
Early layers process low-level visual features that change gradually, while deeper layers encode task-relevant representations that shift more abruptly at instruction transitions; a uniform reuse ratio across layers misallocates computation.
We modulate the per-layer reuse budget via an entropy-based uncertainty proxy~$H_t^{\ell}$ computed from the layer's attention distribution:
\begin{equation}
\rho_t^{\ell}=\mathrm{clip}\!\left(\rho_{\max}-\alpha\, H_t^{\ell},\;\rho_{\min},\;\rho_{\max}\right),
\label{eq:layer-adaptive-reuse}
\end{equation}
where $\alpha$ is a sensitivity coefficient.
High-entropy layers receive a smaller reuse budget; low-entropy layers approach $\rho_{\max}$.
$H_t^\ell$ is read from the existing attention softmax at no extra cost.
When attention weights are not exposed, we apply a global cap $\rho_t \le \rho_{\max}$ uniformly.

\section{\textbf{System Implementation of \textit{VLN-Cache}}}
\label{sec:implementation}


\subsection{\textbf{Cross-Frame Visual Token Caching}}

VLN-Cache accelerates action prediction at each navigation step by reusing the cached key-value (KV) representations of stable visual tokens across consecutive frames, rather than recomputing all $M$ tokens from scratch.
At each timestep $t$, the dual-aware mask $m_t$ (Eq.~\ref{eq:reuse-mask}) partitions all visual tokens into a reuse set $\mathcal{P}_\text{reuse}$ and a fresh-compute set $\mathcal{P}_\text{new}$.
For tokens in $\mathcal{P}_\text{reuse}$, their cached representations are spliced directly into the current decoder state from the view-aligned position $\pi_t(i)$ in the preceding frame (Eq.~\ref{eq:token-splice}); tokens in $\mathcal{P}_\text{new}$ are projected via $W_K$ and $W_V$ as usual.
The key to correct reuse under viewpoint shift lies in the retrieval index $\pi_t(i)$: VLN-Cache traces each candidate token back to the scene surface it observed via the depth-derived remapping (Eq.~\ref{eq:view-remap}), ensuring the retrieved representation always corresponds to the same physical surface regardless of camera displacement.
Because this view-aligned remap preserves geometric correspondence, stable token states are reused across frames, while position-dependent components (e.g., rotary position embeddings) require recomputation at shifted positions; only tokens in $\mathcal{P}_\text{new}$ receive updated RoPE encodings at timestep $t$.
Since no model weights are modified, VLN-Cache operates as a plug-and-play acceleration wrapper compatible with any transformer-based VLA backbone.

\subsection{\textbf{Theoretical Analysis of Computational Complexity}}

The added cost of VLN-Cache is confined to the dual-aware mask computation, which involves a per-token cosine similarity check, a depth-guided neighbourhood search of radius $k$, and a semantic gate aggregation over $L_q$ language query tokens.
The total selection overhead is $\mathcal{O}\!\left(M(D + k^2 + L_q D)\right)$, which is independent of the number of decoder layers $L$ and remains negligible relative to the savings it enables.
The dominant cost in VLA inference is the $W_K$/$W_V$ projection of all $M$ visual tokens repeated across all $L$ decoder layers at every step.
With a reuse ratio $\rho$, only $(1{-}\rho)M$ tokens require fresh projection, yielding a per-step FLOP reduction of approximately $4\rho LMDd_\text{kv}$, where $d_\text{kv}$ is the K/V projection dimension.
For InternVLA-N1 ($L{=}28$, $M{=}196$, $D{=}3584$, $d_\text{kv}{=}512$) at the observed average reuse ratio $\rho{\approx}0.31$ on R2R, this amounts to approximately 12.3\, GFLOPs saved per navigation step (18.3\% of the total attention FLOPs), while the configured upper bound is $\rho_\text{max}{=}0.90$.
All cached tensors are stored on-device in GPU HBM, adding approximately 11.2\,MB per frame (0.027\% of A100 VRAM), so no CPU offloading or PCIe data transfer is required.

\begin{figure}[!t]
    \centering
    \includegraphics[width=\columnwidth]{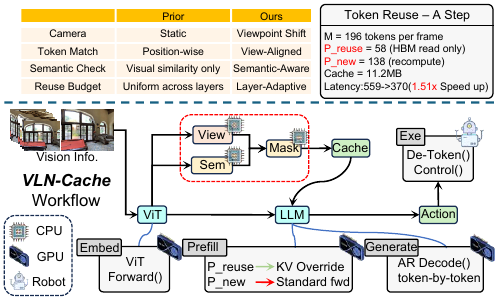} 
    \caption{System implementation pipeline of \textit{VLN-Cache}.}
    \vspace{-4mm}
    \label{fig:workflow}
\end{figure}

\begin{figure*}[!t]
    \centering
    \includegraphics[width=\textwidth]{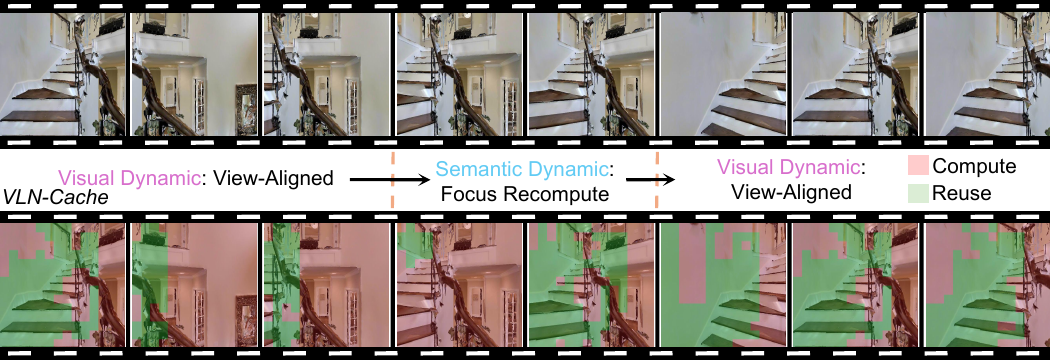}
    \caption{Token reuse visualization on a staircase-to-bedroom navigation episode from R2R-CE.}
    \vspace{-4mm}
    \label{fig:qual-a}
\end{figure*}

\subsection{\textbf{Overall Computation Flow}}

The inference pipeline of VLN-Cache at each navigation step proceeds as follows.
The current RGB-D observation is first encoded by the visual encoder to produce visual token features $\mathbf{H}_t$.
The dual-aware mask then evaluates each token against both the visual similarity criterion and the semantic gate, producing the reuse set $\mathcal{P}_\text{reuse}$.
For tokens in $\mathcal{P}_\text{reuse}$, the view-aligned remap $\pi_t(i)$ locates the corresponding cached entry from the previous step and splices it directly into the decoder state; for tokens in $\mathcal{P}_\text{new}$, standard projections are performed.
The assembled decoder state is then consumed by the language decoder to generate the next navigation action autoregressively.
After decoding, only the non-reused positions in the on-device cache are updated with newly computed states, so the cache remains current for the next step at minimal write cost.
The FLOP savings scale linearly with $\rho$ and compound across the full navigation trajectory; on R2R, the observed token reuse ratio averages approximately $31\%$ across navigation steps (Sec.~\ref{sec:experiments}), translating directly into the measured speedup reported in the experiments.

\section{\textbf{Experiments}}
\label{sec:experiments}

\subsection{\textbf{Setup}}


We apply \textit{VLN-Cache} to InternVLA-N1~\cite{Intern-n1}, a 7B-parameter VLA built on Qwen2.5-VL~\cite{qwen25vl}, and evaluate on the R2R dataset~\cite{r2r} in the VLN-CE continuous environment simulator~\cite{vlnce} using the \texttt{val\_unseen} split of 1{,}839 episodes.
\textit{VLN-Cache} is training-free and requires no fine-tuning or architectural modification.
We compare against a no-cache baseline that runs standard InternVLA-N1 inference with all caching disabled, representing full per-step recomputation.
All experiments run on a single NVIDIA A100-PCIE-40GB GPU (BF16 precision, SDPA attention) paired with an Intel Xeon Silver 4410T CPU.
We report Success Rate (SR) and Success weighted by Path Length (SPL) as navigation metrics, and per-step latency, control frequency, and episode-level wall-clock speedup as efficiency metrics.

\subsection{\textbf{Main Results}}


\textbf{Navigation Accuracy.}\quad
Table~\ref{tab:main-accuracy} compares \textit{VLN-Cache} against representative VLN methods on R2R-CE \texttt{val\_unseen}.
The unmodified DualVLN backbone (InternVLA-N1~\cite{Intern-n1}) achieves SR~=~64.3 and SPL~=~58.5, outperforming all prior methods in this comparison by a substantial margin.
\textit{VLN-Cache} is applied on top of this backbone without any re-training.
By reusing only tokens that are both geometrically and semantically stable across consecutive frames, \textit{VLN-Cache} preserves the visual grounding fidelity of the underlying model, achieving SR~=~63.1 and SPL~=~57.6 with negligible degradation ($\Delta$SR~=~$-1.2\%$).
\begin{table}[!b]
    \vspace{-5mm}
    \centering
    \caption{Comparison with VLN methods on R2R-CE.}
    \vskip -0.05 in
    \label{tab:main-accuracy}
    \resizebox{\columnwidth}{!}{
    \begin{tabular}{l|c|cccc}
    \toprule
    \toprule
    \textbf{Method} & \textbf{Backbone} & \textbf{NE}$\downarrow$ & \textbf{OS}$\uparrow$ & \textbf{SR}$\uparrow$ & \textbf{SPL}$\uparrow$ \\
    \midrule
    \cellcolor{gray!20}{NaVid~\cite{navid}} & \cellcolor{gray!20}{Vicuna-7B} & \cellcolor{gray!20}{5.47} & \cellcolor{gray!20}{49.1} & \cellcolor{gray!20}{37.4} & \cellcolor{gray!20}{35.9} \\
    \cellcolor{teal!15}{MapNav~\cite{mapnav}} & \cellcolor{teal!15}{Qwen2-7B} & \cellcolor{teal!15}{4.93} & \cellcolor{teal!15}{53.0} & \cellcolor{teal!15}{39.7} & \cellcolor{teal!15}{37.2} \\
    \cellcolor{cyan!15}{UniNaVid~\cite{Uni-NaVid}} & \cellcolor{cyan!15}{Vicuna-7B} & \cellcolor{cyan!15}{5.58} & \cellcolor{cyan!15}{53.3} & \cellcolor{cyan!15}{47.0} & \cellcolor{cyan!15}{42.7} \\
    \cellcolor{pink!20}{NaVILA~\cite{NaVILA}} & \cellcolor{pink!20}{Llama-3-8B} & \cellcolor{pink!20}{5.22} & \cellcolor{pink!20}{62.5} & \cellcolor{pink!20}{54.0} & \cellcolor{pink!20}{49.0} \\
    \cellcolor{orange!20}{StreamVLN~\cite{StreamVLN}} & \cellcolor{orange!20}{Qwen2-7B} & \cellcolor{orange!20}{4.98} & \cellcolor{orange!20}{64.2} & \cellcolor{orange!20}{56.9} & \cellcolor{orange!20}{51.9} \\
    \cellcolor{yellow!20}{DualVLN~\cite{Intern-n1}} & \cellcolor{yellow!20}{Qwen2.5-VL-7B} & \cellcolor{yellow!20}{4.05} & \cellcolor{yellow!20}{70.7} & \cellcolor{yellow!20}{64.3} & \cellcolor{yellow!20}{58.5} \\
    \cellcolor{green!20}{\textbf{\textit{VLN-Cache} (Ours)}} & \cellcolor{green!20}{Qwen2.5-VL-7B} & \cellcolor{green!20}{\textbf{3.93}} & \cellcolor{green!20}{\textbf{71.4}} & \cellcolor{green!20}{\textbf{63.1}} & \cellcolor{green!20}{\textbf{57.6}} \\
    \bottomrule
    \bottomrule
    \end{tabular}
    }
\end{table}
w

\textbf{Inference Efficiency.}\quad
Table~\ref{tab:main-efficiency} reports efficiency measurements from a 10-episode pilot run.
Without caching, the System-2 VLA planner incurs a per-step latency of 637~ms; with \textit{VLN-Cache}, latency reduces to 419~ms, yielding a \textbf{1.52$\times$ step-level speedup}.
Because per-step latency is reduced, the accumulated wall-clock time over an episode decreases: wall-clock episode time decreases from 114.7~s to 75.5~s, achieving a \textbf{1.52$\times$ episode-level speedup}.
On average \textbf{31\%} of per-layer VLA tokens are reused from the cache per step, with higher reuse during straight cruising segments and lower reuse during sharp turns, confirming that the dual-aware design correctly adapts to both visual and semantic dynamics along the trajectory.
The per-frame cache footprint is approximately 11.2~MB (0.027\% of A100 VRAM), imposing negligible memory overhead.

\subsection{\textbf{Qualitative Analysis}}

\begin{figure*}[!t]
    \centering
    \includegraphics[width=\textwidth]{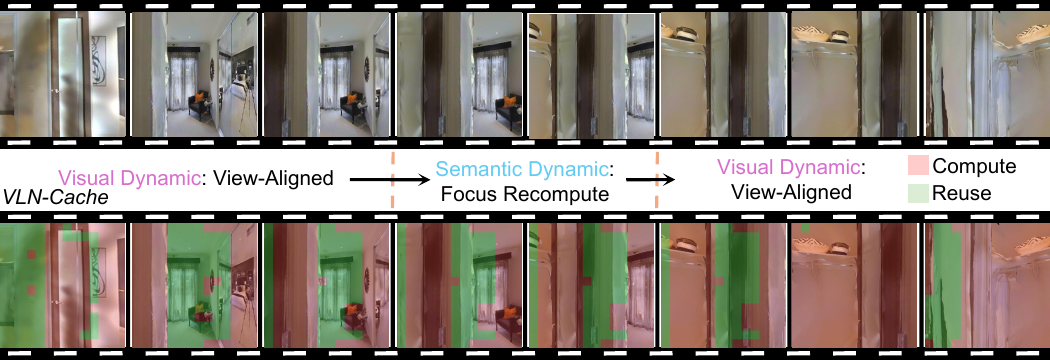}
    \caption{Token reuse visualization on a bedroom-exit-to-rug navigation episode from R2R-CE.}
    \vspace{-4mm}
    \label{fig:qual-b}
\end{figure*}

Figs.~\ref{fig:qual-a} and~\ref{fig:qual-b} show two representative R2R-CE episodes.
In Episode~1 (staircase $\to$ bedroom $\to$ closet), large viewpoint shifts at turns drive the reuse rate down, while straight segments toward the closet sustain high reuse—the agent's trajectory remains indistinguishable from the no-cache baseline.
In Episode~2 (bedroom exit $\to$ straight corridor $\to$ rug), the long straight segment yields persistently high background reuse; the semantic gate forces fresh computation for the couch and rug landmarks upon approach, preserving correct goal recognition.
Across both episodes, \textit{VLN-Cache} concentrates computation on viewpoint-shifted and semantically critical tokens without affecting route fidelity.

\subsection{\textbf{Ablation Study}}


\begin{table}[!b]
    \vspace{-3mm}
    \centering
    \caption{Inference efficiency comparison on R2R-CE.}
    \vskip -0.05 in
    \label{tab:main-efficiency}
    \resizebox{\columnwidth}{!}{
    \begin{tabular}{l|c|cc|c}
    \toprule
    \toprule
    \textbf{Method} & \textbf{Latency}$\downarrow$ & \textbf{Step} & \textbf{Episode} & \textbf{Token} \\
     & \textbf{(ms/step)} & \textbf{Speedup} & \textbf{Speedup} & \textbf{Reuse} \\
    \midrule
    \cellcolor{blue!15}{DualVLN~\cite{Intern-n1}} & \cellcolor{blue!15}{637} & \cellcolor{blue!15}{1.00$\times$} & \cellcolor{blue!15}{1.00$\times$} & \cellcolor{blue!15}{0\%} \\  
    \cellcolor{green!20}{\textbf{\textit{VLN-Cache} (Ours)}} & \cellcolor{green!20}{\textbf{419}} & \cellcolor{green!20}{\textbf{1.52$\times$}} & \cellcolor{green!20}{\textbf{1.52$\times$}} & \cellcolor{green!20}{\textbf{31\%}} \\
    \bottomrule
    \bottomrule
    \end{tabular}
    }
    
\end{table}

We conduct ablation studies on R2R-CE \texttt{val\_unseen} to evaluate the contribution of each component, with results reported in Table~\ref{tab:ablation}.

When view-aligned remap is removed and cache reuse falls back to the position-wise scheme (i.e., $\pi_t(i) = i$, equivalent to applying VLA-Cache~\cite{vlacache} directly to the VLN setting), SR and SPL drop sharply despite a comparable token reuse ratio.
This confirms the central argument of Section~\ref{sec:analysis}: under continuous viewpoint shift, cached tokens at position $i$ no longer correspond to the same scene patch at time $t$, and injecting geometrically misaligned representations actively degrades the language decoder's grounding ability.

Building on view-aligned remap, we further examine each gate in isolation.
Without the semantic gate, the system relies solely on visual cosine similarity to determine reuse.
Navigation accuracy degrades because appearance-based filtering is blind to instruction-driven semantic shifts: tokens belonging to landmarks that have become irrelevant to the current sub-goal remain marked as reusable by appearance alone, silently corrupting the planner's contextual reasoning.
Conversely, without the visual gate, the system depends entirely on the semantic gate.
Navigation accuracy falls below the no-cache baseline (SR~=~61.7 vs.\ 64.3), as visually changed tokens are incorrectly reused, introducing feature drift; the configuration does retain a higher speedup.

\begin{figure}[!t]
    \centering
    \includegraphics[width=\columnwidth]{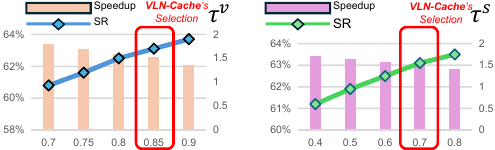}
    \caption{Sensitivity to $\tau_v$  and $\tau_s$.}
    \vspace{-4mm}
    \label{fig:hyper-1}
\end{figure}

\begin{figure}[!t]
    \centering
    \includegraphics[width=\columnwidth]{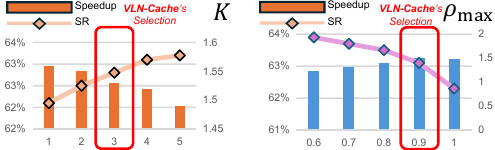}
    \caption{Sensitivity to $k$  and $\rho_\text{max}$.}
    \vspace{-4mm}
    \label{fig:hyper-2}
\end{figure}

The complete \textit{VLN-Cache}, combining view-aligned remap with both gates, achieves the best accuracy-efficiency trade-off, recovering near-baseline SR/SPL while maintaining meaningful speedup.
The two dynamics identified in Section~\ref{sec:analysis}, visual and semantic, require orthogonal remedies that compose without interference, and the ablation results confirm that each component provides an independent and non-redundant contribution to the overall system performance.

\subsection{\textbf{Discussion}}

\begin{table}[!b]
    \centering
    \vspace{-5mm}
    \caption{Ablation study on R2R-CE.}
    \vskip -0.05 in
    \label{tab:ablation}
    \resizebox{\columnwidth}{!}{
    \begin{tabular}{l|cc|cc}
    \toprule
    \toprule
    \textbf{Configuration} & \textbf{SR}$\uparrow$ & \textbf{SPL}$\uparrow$ & \textbf{Step Sp.} & \textbf{Token Reuse} \\
    \midrule
    \cellcolor{gray!20}{w/o \textit{VLN-Cache} (Baseline)} & \cellcolor{gray!20}{64.3} & \cellcolor{gray!20}{58.5} & \cellcolor{gray!20}{1.00$\times$} & \cellcolor{gray!20}{0\%} \\
    \cellcolor{pink!20}{w/o View-Aligned Remap} & \cellcolor{pink!20}{62.4} & \cellcolor{pink!20}{56.8} & \cellcolor{pink!20}{1.41$\times$} & \cellcolor{pink!20}{43\%} \\
    \cellcolor{orange!20}{w/o Semantic Gate} & \cellcolor{orange!20}{62.9} & \cellcolor{orange!20}{57.3} & \cellcolor{orange!20}{1.47$\times$} & \cellcolor{orange!20}{30\%} \\
    \cellcolor{yellow!20}{w/o Visual Gate} & \cellcolor{yellow!20}{61.7} & \cellcolor{yellow!20}{56.1} & \cellcolor{yellow!20}{1.61$\times$} & \cellcolor{yellow!20}{49\%} \\
    \cellcolor{green!20}{\textbf{Full \textit{VLN-Cache} (Ours)}} & \cellcolor{green!20}{\textbf{63.1}} & \cellcolor{green!20}{\textbf{57.6}} & \cellcolor{green!20}{\textbf{1.52$\times$}} & \cellcolor{green!20}{\textbf{31\%}} \\
    \bottomrule
    \bottomrule
    \end{tabular}
    }
\end{table}


\noindent \textbf{Hyper-Parameters.}
\textit{VLN-Cache} involves four hyper-parameters: $\tau_v$, $\tau_s$, $k$, and $\rho_\text{max}$.
We evaluate \textit{VLN-Cache}'s performance across various hyper-parameter values on the R2R-CE benchmark, with results presented in Figs.~\ref{fig:hyper-1} and~\ref{fig:hyper-2}.
As illustrated in Fig.~\ref{fig:hyper-1}, we find that $\tau_v$ exerts a significant influence on SR and speedup, as it controls how aggressively the Visual Gate accepts tokens for reuse.
Considering both SR and speedup comprehensively, we select $\tau_v = 0.85$ as the default.
Similarly, $\tau_s$ also influences SR and speedup, as it governs the frequency of Semantic Gate refreshes; we select $\tau_s = 0.70$.
For $k$, setting $k = 0$ degenerates to position-wise reuse and incurs the accuracy collapse shown in the ablation; we default to $k = 3$.
$\rho_\text{max}$ and the remaining parameters $\Delta_s$, $\epsilon$ exhibit low sensitivity across the tested range and are fixed at $0.90$, $0.30$, and $10^{-6}$ respectively.

\noindent \textbf{Generality.}
\textit{VLN-Cache} is training-free and imposes no specific requirements on VLA backbone architecture, navigation task category, or simulation environment.
The framework exhibits good generality and is directly applicable to any autoregressive VLA model used as a System-2 planner in dual-system VLN frameworks.

\noindent \textbf{Scope.}
We do not design an automatic hyper-parameter determination method, mainly because established standards for continuous VLN environments are currently lacking.
Automatic hyper-parameter selection is therefore designated as future work.
Additionally, the current implementation assumes access to depth and relative pose from the simulator or odometry for view-aligned remapping, and does not address map-based navigation settings.

\section{\textbf{Conclusion}}
\label{sec:conclusion}

In this paper, we first characterize two independent dynamics in VLN inference that jointly undermine position-wise token caching: viewpoint-induced geometric misalignment and instruction-driven semantic staleness.
Based on these insights, we propose \textit{VLN-Cache}, a dual-aware token caching framework with view-aligned remap, dual gating, and entropy-based layer-adaptive reuse.
\textit{VLN-Cache} is training-free and integrates directly into the System-2 VLA planner of dual-system VLN architectures without any architectural modification.
Experiments on R2R-CE demonstrate that \textit{VLN-Cache} achieves a $1.52\times$ step-level speedup and $1.52\times$ episode-level speedup with negligible navigation accuracy degradation (SR drop $<1.5\%$).
We hope this work motivates further exploration of inference efficiency for VLN agents in continuous environments.









\bibliographystyle{iros2026}
\bibliography{ref/reference.bib}

@inproceedings{VLN_servey1,
   title={Vision-and-Language Navigation: A Survey of Tasks, Methods, and Future Directions},
   url={http://dx.doi.org/10.18653/v1/2022.acl-long.524},
   DOI={10.18653/v1/2022.acl-long.524},
   booktitle={Proceedings of the 60th Annual Meeting of the Association for Computational Linguistics (Volume 1: Long Papers)},
   publisher={Association for Computational Linguistics},
   author={Gu, Jing and Stefani, Eliana and Wu, Qi and Thomason, Jesse and Wang, Xin},
   year={2022},
   pages={7606–7623} 
}

@misc{VLN_servey2,
      title={Vision-and-Language Navigation: Interpreting visually-grounded navigation instructions in real environments}, 
      author={Peter Anderson and Qi Wu and Damien Teney and Jake Bruce and Mark Johnson and Niko Sünderhauf and Ian Reid and Stephen Gould and Anton van den Hengel},
      year={2018},
      eprint={1711.07280},
      archivePrefix={arXiv},
      primaryClass={cs.CV},
      url={https://arxiv.org/abs/1711.07280}, 
}

@misc{VL_Nav,
      title={VL-Nav: Real-time Vision-Language Navigation with Spatial Reasoning}, 
      author={Yi Du and Taimeng Fu and Zhuoqun Chen and Bowen Li and Shaoshu Su and Zhipeng Zhao and Chen Wang},
      year={2025},
      eprint={2502.00931},
      archivePrefix={arXiv},
      primaryClass={cs.RO},
      url={https://arxiv.org/abs/2502.00931}, 
}

@misc{IROS_VLN,
      title={IROS: A Dual-Process Architecture for Real-Time VLM-Based Indoor Navigation}, 
      author={Joonhee Lee and Hyunseung Shin and Jeonggil Ko},
      year={2026},
      eprint={2601.21506},
      archivePrefix={arXiv},
      primaryClass={cs.RO},
      url={https://arxiv.org/abs/2601.21506}, 
}

@misc{HIAI,
      title={Harnessing Input-Adaptive Inference for Efficient VLN}, 
      author={Dongwoo Kang and Akhil Perincherry and Zachary Coalson and Aiden Gabriel and Stefan Lee and Sanghyun Hong},
      year={2025},
      eprint={2508.09262},
      archivePrefix={arXiv},
      primaryClass={cs.CV},
      url={https://arxiv.org/abs/2508.09262}, 
}

@misc{efficientvln,
      title={Efficient-VLN: A Training-Efficient Vision-Language Navigation Model}, 
      author={Duo Zheng and Shijia Huang and Yanyang Li and Liwei Wang},
      year={2025},
      eprint={2512.10310},
      archivePrefix={arXiv},
      primaryClass={cs.CV},
      url={https://arxiv.org/abs/2512.10310}, 
}

@misc{etp_r1,
      title={ETP-R1: Evolving Topological Planning with Reinforcement Fine-tuning for Vision-Language Navigation in Continuous Environments}, 
      author={Shuhao Ye and Sitong Mao and Yuxiang Cui and Xuan Yu and Shichao Zhai and Wen Chen and Shunbo Zhou and Rong Xiong and Yue Wang},
      year={2025},
      eprint={2512.20940},
      archivePrefix={arXiv},
      primaryClass={cs.RO},
      url={https://arxiv.org/abs/2512.20940}, 
}

@misc{StreamVLN,
      title={StreamVLN: Streaming Vision-and-Language Navigation via SlowFast Context Modeling}, 
      author={Meng Wei and Chenyang Wan and Xiqian Yu and Tai Wang and Yuqiang Yang and Xiaohan Mao and Chenming Zhu and Wenzhe Cai and Hanqing Wang and Yilun Chen and Xihui Liu and Jiangmiao Pang},
      year={2025},
      eprint={2507.05240},
      archivePrefix={arXiv},
      primaryClass={cs.RO},
      url={https://arxiv.org/abs/2507.05240}, 
}

@misc{MiniVLN,
      title={MiniVLN: Efficient Vision-and-Language Navigation by Progressive Knowledge Distillation}, 
      author={Junyou Zhu and Yanyuan Qiao and Siqi Zhang and Xingjian He and Qi Wu and Jing Liu},
      year={2024},
      eprint={2409.18800},
      archivePrefix={arXiv},
      primaryClass={cs.CV},
      url={https://arxiv.org/abs/2409.18800}, 
}

@misc{NaVILA,
      title={NaVILA: Legged Robot Vision-Language-Action Model for Navigation}, 
      author={An-Chieh Cheng and Yandong Ji and Zhaojing Yang and Zaitian Gongye and Xueyan Zou and Jan Kautz and Erdem Bıyık and Hongxu Yin and Sifei Liu and Xiaolong Wang},
      year={2025},
      eprint={2412.04453},
      archivePrefix={arXiv},
      primaryClass={cs.RO},
      url={https://arxiv.org/abs/2412.04453}, 
}

@misc{Uni-NaVid,
      title={Uni-NaVid: A Video-based Vision-Language-Action Model for Unifying Embodied Navigation Tasks}, 
      author={Jiazhao Zhang and Kunyu Wang and Shaoan Wang and Minghan Li and Haoran Liu and Songlin Wei and Zhongyuan Wang and Zhizheng Zhang and He Wang},
      year={2025},
      eprint={2412.06224},
      archivePrefix={arXiv},
      primaryClass={cs.RO},
      url={https://arxiv.org/abs/2412.06224}, 
}

@misc{NAPVLN,
      title={Walk and Read Less: Improving the Efficiency of Vision-and-Language Navigation via Tuning-Free Multimodal Token Pruning}, 
      author={Wenda Qin and Andrea Burns and Bryan A. Plummer and Margrit Betke},
      year={2025},
      eprint={2509.15250},
      archivePrefix={arXiv},
      primaryClass={cs.CV},
      url={https://arxiv.org/abs/2509.15250}, 
}

@misc{vlacache,
      title={VLA-Cache: Efficient Vision-Language-Action Manipulation via Adaptive Token Caching}, 
      author={Siyu Xu and Yunke Wang and Chenghao Xia and Dihao Zhu and Tao Huang and Chang Xu},
      year={2025},
      eprint={2502.02175},
      archivePrefix={arXiv},
      primaryClass={cs.RO},
      url={https://arxiv.org/abs/2502.02175}, 
}

@misc{token_merging,
      title={Token Merging: Your ViT But Faster}, 
      author={Daniel Bolya and Cheng-Yang Fu and Xiaoliang Dai and Peizhao Zhang and Christoph Feichtenhofer and Judy Hoffman},
      year={2023},
      eprint={2210.09461},
      archivePrefix={arXiv},
      primaryClass={cs.CV},
      url={https://arxiv.org/abs/2210.09461}, 
}

@misc{stcache,
      title={Prune Spatio-temporal Tokens by Semantic-aware Temporal Accumulation}, 
      author={Shuangrui Ding and Peisen Zhao and Xiaopeng Zhang and Rui Qian and Hongkai Xiong and Qi Tian},
      year={2023},
      eprint={2308.04549},
      archivePrefix={arXiv},
      primaryClass={cs.CV},
      url={https://arxiv.org/abs/2308.04549}, 
}

@misc{token_sparsification,
      title={Making Vision Transformers Efficient from A Token Sparsification View}, 
      author={Shuning Chang and Pichao Wang and Ming Lin and Fan Wang and David Junhao Zhang and Rong Jin and Mike Zheng Shou},
      year={2023},
      eprint={2303.08685},
      archivePrefix={arXiv},
      primaryClass={cs.CV},
      url={https://arxiv.org/abs/2303.08685}, 
}

@misc{EgoPrune,
      title={EgoPrune: Efficient Token Pruning for Egomotion Video Reasoning in Embodied Agent}, 
      author={Jiaao Li and Kaiyuan Li and Chen Gao and Yong Li and Xinlei Chen},
      year={2025},
      eprint={2507.15428},
      archivePrefix={arXiv},
      primaryClass={cs.CV},
      url={https://arxiv.org/abs/2507.15428}, 
}

@misc{View_Invariant_Learning,
      title={View Invariant Learning for Vision-Language Navigation in Continuous Environments}, 
      author={Josh Qixuan Sun and Xiaoying Xing and Huaiyuan Weng and Chul Min Yeum and Mark Crowley},
      year={2025},
      eprint={2507.08831},
      archivePrefix={arXiv},
      primaryClass={cs.CV},
      url={https://arxiv.org/abs/2507.08831}, 
}

@misc{Learn_VLA_cache,
      title={Learning to Accelerate Vision-Language-Action Models through Adaptive Visual Token Caching}, 
      author={Yujie Wei and Jiahan Fan and Jiyu Guo and Ruichen Zhen and Rui Shao and Xiu Su and Zeke Xie and Shuo Yang},
      year={2026},
      eprint={2602.00686},
      archivePrefix={arXiv},
      primaryClass={cs.RO},
      url={https://arxiv.org/abs/2602.00686}, 
}

@misc{Intern-n1,
      title={Ground Slow, Move Fast: A Dual-System Foundation Model for Generalizable Vision-and-Language Navigation}, 
      author={Meng Wei and Chenyang Wan and Jiaqi Peng and Xiqian Yu and Yuqiang Yang and Delin Feng and Wenzhe Cai and Chenming Zhu and Tai Wang and Jiangmiao Pang and Xihui Liu},
      year={2025},
      eprint={2512.08186},
      archivePrefix={arXiv},
      primaryClass={cs.RO},
      url={https://arxiv.org/abs/2512.08186}, 
}

@misc{qwen25vl,
      title={qwen25vl}, 
      author={Shuai Bai and Keqin Chen and Xuejing Liu and Jialin Wang and Wenbin Ge and Sibo Song and Kai Dang and Peng Wang and Shijie Wang and Jun Tang and Humen Zhong and Yuanzhi Zhu and Mingkun Yang and Zhaohai Li and Jianqiang Wan and Pengfei Wang and Wei Ding and Zheren Fu and Yiheng Xu and Jiabo Ye and Xi Zhang and Tianbao Xie and Zesen Cheng and Hang Zhang and Zhibo Yang and Haiyang Xu and Junyang Lin},
      year={2025},
      eprint={2502.13923},
      archivePrefix={arXiv},
      primaryClass={cs.CV},
      url={https://arxiv.org/abs/2502.13923}, 
}

@misc{r2r,
      title={Vision-and-Language Navigation: Interpreting visually-grounded navigation instructions in real environments}, 
      author={Peter Anderson and Qi Wu and Damien Teney and Jake Bruce and Mark Johnson and Niko Sünderhauf and Ian Reid and Stephen Gould and Anton van den Hengel},
      year={2018},
      eprint={1711.07280},
      archivePrefix={arXiv},
      primaryClass={cs.CV},
      url={https://arxiv.org/abs/1711.07280}, 
}

@misc{vlnce,
      title={Beyond the Nav-Graph: Vision-and-Language Navigation in Continuous Environments}, 
      author={Jacob Krantz and Erik Wijmans and Arjun Majumdar and Dhruv Batra and Stefan Lee},
      year={2020},
      eprint={2004.02857},
      archivePrefix={arXiv},
      primaryClass={cs.CV},
      url={https://arxiv.org/abs/2004.02857}, 
}

@misc{navid,
      title={NaVid: Video-based VLM Plans the Next Step for Vision-and-Language Navigation}, 
      author={Jiazhao Zhang and Kunyu Wang and Rongtao Xu and Gengze Zhou and Yicong Hong and Xiaomeng Fang and Qi Wu and Zhizheng Zhang and He Wang},
      year={2024},
      eprint={2402.15852},
      archivePrefix={arXiv},
      primaryClass={cs.CV},
      url={https://arxiv.org/abs/2402.15852}, 
}

@misc{mapnav,
      title={MapNav: A Novel Memory Representation via Annotated Semantic Maps for Vision-and-Language Navigation}, 
      author={Lingfeng Zhang and Xiaoshuai Hao and Qinwen Xu and Qiang Zhang and Xinyao Zhang and Pengwei Wang and Jing Zhang and Zhongyuan Wang and Shanghang Zhang and Renjing Xu},
      year={2025},
      eprint={2502.13451},
      archivePrefix={arXiv},
      primaryClass={cs.RO},
      url={https://arxiv.org/abs/2502.13451}, 
}

@misc{KERV,
      title={KERV: Kinematic-Rectified Speculative Decoding for Embodied VLA Models}, 
      author={Zihao Zheng and Zhihao Mao and Maoliang Li and Jiayu Chen and Xinhao Sun and Zhaobo Zhang and Donggang Cao and Hong Mei and Xiang Chen},
      year={2026},
      eprint={2603.01581},
      archivePrefix={arXiv},
      primaryClass={cs.RO},
      url={https://arxiv.org/abs/2603.01581}, 
}

@misc{RT2,
      title={RT-2: Vision-Language-Action Models Transfer Web Knowledge to Robotic Control}, 
      author={Anthony Brohan and Noah Brown and Justice Carbajal and Yevgen Chebotar and Xi Chen and Krzysztof Choromanski and Tianli Ding and Danny Driess and Avinava Dubey and Chelsea Finn and Pete Florence and Chuyuan Fu and Montse Gonzalez Arenas and Keerthana Gopalakrishnan and Kehang Han and Karol Hausman and Alexander Herzog and Jasmine Hsu and Brian Ichter and Alex Irpan and Nikhil Joshi and Ryan Julian and Dmitry Kalashnikov and Yuheng Kuang and Isabel Leal and Lisa Lee and Tsang-Wei Edward Lee and Sergey Levine and Yao Lu and Henryk Michalewski and Igor Mordatch and Karl Pertsch and Kanishka Rao and Krista Reymann and Michael Ryoo and Grecia Salazar and Pannag Sanketi and Pierre Sermanet and Jaspiar Singh and Anikait Singh and Radu Soricut and Huong Tran and Vincent Vanhoucke and Quan Vuong and Ayzaan Wahid and Stefan Welker and Paul Wohlhart and Jialin Wu and Fei Xia and Ted Xiao and Peng Xu and Sichun Xu and Tianhe Yu and Brianna Zitkovich},
      year={2023},
      eprint={2307.15818},
      archivePrefix={arXiv},
      primaryClass={cs.RO},
      url={https://arxiv.org/abs/2307.15818}, 
}

@misc{FastV,
      title={An Image is Worth 1/2 Tokens After Layer 2: Plug-and-Play Inference Acceleration for Large Vision-Language Models}, 
      author={Liang Chen and Haozhe Zhao and Tianyu Liu and Shuai Bai and Junyang Lin and Chang Zhou and Baobao Chang},
      year={2024},
      eprint={2403.06764},
      archivePrefix={arXiv},
      primaryClass={cs.CV},
      url={https://arxiv.org/abs/2403.06764}, 
}

@misc{LLaVA_PruMerge,
      title={LLaVA-PruMerge: Adaptive Token Reduction for Efficient Large Multimodal Models}, 
      author={Yuzhang Shang and Mu Cai and Bingxin Xu and Yong Jae Lee and Yan Yan},
      year={2026},
      eprint={2403.15388},
      archivePrefix={arXiv},
      primaryClass={cs.CV},
      url={https://arxiv.org/abs/2403.15388}, 
}

@misc{SnapKV,
      title={SnapKV: LLM Knows What You are Looking for Before Generation}, 
      author={Yuhong Li and Yingbing Huang and Bowen Yang and Bharat Venkitesh and Acyr Locatelli and Hanchen Ye and Tianle Cai and Patrick Lewis and Deming Chen},
      year={2024},
      eprint={2404.14469},
      archivePrefix={arXiv},
      primaryClass={cs.CL},
      url={https://arxiv.org/abs/2404.14469}, 
}

@misc{OpenVLA,
      title={OpenVLA: An Open-Source Vision-Language-Action Model}, 
      author={Moo Jin Kim and Karl Pertsch and Siddharth Karamcheti and Ted Xiao and Ashwin Balakrishna and Suraj Nair and Rafael Rafailov and Ethan Foster and Grace Lam and Pannag Sanketi and Quan Vuong and Thomas Kollar and Benjamin Burchfiel and Russ Tedrake and Dorsa Sadigh and Sergey Levine and Percy Liang and Chelsea Finn},
      year={2024},
      eprint={2406.09246},
      archivePrefix={arXiv},
      primaryClass={cs.RO},
      url={https://arxiv.org/abs/2406.09246}, 
}

\end{document}